\pdfoutput=1

\documentclass[runningheads]{llncs}
\usepackage{graphicx}

\usepackage{tikz}
\usepackage{comment}
\usepackage{amsmath,amssymb} 
\usepackage{color}

\usepackage[ruled]{algorithm2e}
\usepackage[noend]{algpseudocode}
\usepackage{wrapfig}
\usepackage{tablefootnote}
\usepackage{paralist}  
\usepackage[export]{adjustbox}
\usepackage{array}
\usepackage{pbox}
\usepackage{multirow}
\usepackage{booktabs}
\usepackage{soul}
\usepackage{caption}
\usepackage{enumitem}
\usepackage{textcomp}
\usepackage{gensymb}
\usepackage{xcolor}
\definecolor{citecolor}{HTML}{0071bc}
\usepackage[pagebackref=true,breaklinks=true,colorlinks=true,citecolor=citecolor,linkcolor=citecolor, bookmarks=false]{hyperref}
\usepackage{subcaption}
\usepackage{xcolor}
\usepackage[normalem]{ulem}

\usepackage{microtype}
\usepackage{colortbl}
\usepackage{cite}
\newcommand{\rpm}{\raisebox{.2ex}{$\scriptstyle\pm$}}
\makeatletter
\DeclareRobustCommand\onedot{\futurelet\@let@token\@onedot}
\def\@onedot{\ifx\@let@token.\else.\null\fi\xspace}

\def\eg{\emph{e.g}\onedot} 
\def\ie{\emph{i.e}\onedot} 
 
\def\etc{\emph{etc}\onedot} 
 
\def\etal{\emph{et al}\onedot}
\def\ds{${\cal D}_s$\xspace}
\def\dss{${\cal D}_{ss}$\xspace}
\makeatother

\newcommand\jcsu[1]{{#1}}

\begin{document}
\pagestyle{headings}
\mainmatter
\def\ECCVSubNumber{352}  

\title{When Does Self-supervision Improve \\ Few-shot Learning?}

%
\newcommand{\orcid}[1]{\href{https://orcid.org/#1}{\protect\includegraphics[width=8pt]{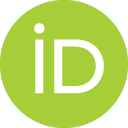}}}
\author{Jong-Chyi Su\inst{1}\orcid{0000-0002-7933-8308} \qquad
Subhransu Maji\inst{1}\orcid{0000-0002-3869-9334} \qquad
Bharath Hariharan\inst{2}\orcid{0000-0002-2309-4703}}

\authorrunning{Su et al.}
\institute{University of Massachusetts Amherst \and Cornell University \\
\email{\{jcsu,smaji\}@cs.umass.edu, bharathh@cs.cornell.edu}
}


\maketitle

\begin{abstract}
  We investigate the role of self-supervised learning (SSL) in the context of few-shot learning. Although recent research has shown the benefits of SSL on large unlabeled datasets, its utility on small datasets is relatively unexplored. We find that SSL reduces the relative error rate of few-shot meta-learners by 4\%-27\%, even when the datasets are small and \emph{only} utilizing images within the datasets. The improvements are greater when the training set is smaller or the task is more challenging. Although the benefits of SSL may increase with larger training sets, we observe that SSL can hurt the performance when the distributions of images used for meta-learning and SSL are different. We conduct a systematic study by varying the degree of domain shift and analyzing the performance of several meta-learners on a multitude of domains. Based on this analysis we present a technique that automatically selects images for SSL from a large, generic pool of unlabeled images for a given dataset that provides further improvements.
\end{abstract}

\section{Introduction}
Current machine learning algorithms require enormous amounts of
training data to learn new tasks.
This is an issue for many practical problems across domains such as
biology and medicine where labeled data is hard to come by.
In contrast, we humans can quickly learn new concepts from limited
training data by relying on our past ``visual experience''.
Recent work attempts to emulate this by training a feature
representation to classify a training dataset of ``base'' classes with the hope that
the resulting representation generalizes not just to unseen examples
of the same classes but also to novel classes, which may have very few training examples (called few-shot learning).
However, training for base class classification can force the network
to only encode features that are useful for distinguishing between
base classes.
In the process, it might discard semantic information that is
irrelevant for base classes but critical for novel classes.
This might be especially true when the base dataset is small or when
the class distinctions are challenging.

One way to recover this useful semantic information is to leverage
representation learning techniques that do not use class labels,
namely, \emph{unsupervised} or \emph{self-supervised learning}.
The key idea is to learn about statistical regularities within images, such as the spatial
relationship between patches, or its orientation, that might be a cue to semantics.
Despite recent advances, these techniques have only been applied to a
few domains (\eg, entry-level classes on internet imagery), and under
the assumption that large amounts of unlabeled images are available.
Their applicability to the general few-shot scenario is unclear.
In particular, can these techniques prevent overfitting to
base classes and improve performance on novel
classes in the few-shot setting?
If so, does the benefit generalize across domains and to more challenging tasks?
Moreover, can self-supervision boost performance in
domains where even unlabeled images are hard to get?

\begin{figure*}[t]
\centering
\includegraphics[width=\linewidth]{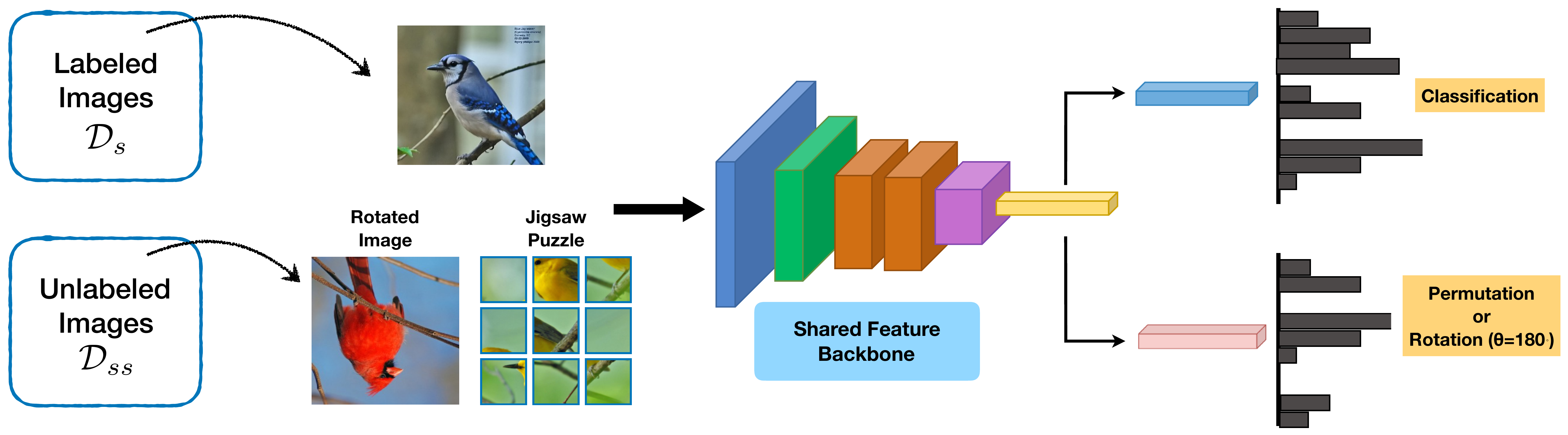}
\caption{\label{fig:overview}
  \textbf{Combining supervised and
  self-supervised losses for few-shot learning.}
  Self-supervised tasks such as jigsaw puzzle or rotation
  prediction act as a
  data-dependent regularizer for the shared feature backbone.
  Our work investigates how the performance on the \emph{target task domain}
  (\ds) is impacted by the choice of the \emph{domain used for
  self-supervision} (\dss).
}
\end{figure*}

This paper seeks to answer these questions.
We show that with \emph{no additional training data}, adding a  
self-supervised task as an auxiliary task (Fig.~\ref{fig:overview}) improves the performance of 
existing few-shot techniques on benchmarks across a multitude of domains
(Fig.~\ref{fig:fsl_result}), in agreement with conclusions from similar recent work~\cite{gidaris2019boosting}.
Intriguingly, we find that the benefits of self-supervision \emph{increase} with the
difficulty of the task, for example when training from a smaller base dataset, or with degraded inputs such as low resolution or greyscale
images (Fig.~\ref{fig:low_quality_result}).

One might surmise that as with traditional SSL, additional unlabeled images might improve performance further.
But what unlabeled images should we use for novel problem domains where unlabeled data is not freely available?
To answer this, we conduct a series of experiments with additional unlabeled data from different domains.
We find that adding more unlabeled images improves performance \emph{only} when the images used for self-supervision are within the \emph{same domain}
as the base classes (Fig.~\ref{fig:vary}); otherwise, they can even \emph{negatively} impact the 
performance of the few-shot learner (Fig.~\ref{fig:mixed}).
Based on this analysis, we present a simple approach that uses a domain classifier to pick similar-domain unlabeled images for self-supervision from a large and generic pool of images (Fig.~\ref{fig:selected}).
The resulting method improves over the 
performance of a model trained with self-supervised learning from images within
the dataset (Fig.~\ref{fig:pool}).
Taken together, this results in a powerful, general, and practical approach for improving few-shot learning on small datasets in novel domains.
Finally, these benefits are also observed in standard classification tasks (Appendix~A.3).

\section{Related Work}

\subsubsection{Few-shot learning}
Few-shot learning aims to learn representations that generalize well
to the novel classes where only a few images are available.
To this end, several meta-learning approaches have been proposed that
evaluate representations by sampling many few-shot tasks within the
domain of a \emph{base} dataset.
These include optimization-based meta-learners, such as
model-agnostic meta-learner (MAML)~\cite{finn2017model}, gradient
unrolling~\cite{ravi2016optimization}, closed-form
solvers~\cite{bertinetto2018meta}, and convex
learners~\cite{lee2019meta}. 
The second class of methods rely on distance-based classifiers such as matching networks~\cite{vinyals2016matching} and
prototypical networks (ProtoNet)~\cite{snell2017prototypical}.
Another class of methods~\cite{gidaris2018dynamic,qi2018low,qiao2018few} model the mapping between training data and classifier weights using a feed-forward network.


While the literature is rapidly growing, a recent
study by Chen~\etal~\cite{chen2019closerfewshot} has shown that the differences between
meta-learners are diminished when deeper networks are used.
They develop a strong baseline for few-shot learning 
and show that the performance of ProtoNet~\cite{snell2017prototypical}
matches or surpasses several recently proposed meta-learners.
We build our experiments on top of this work and show that auxiliary
self-supervised tasks provide additional benefits across a large array
of few-shot benchmarks and across meta-learners.


\subsubsection{Self-supervised learning}
Human labels are expensive to collect and hard to scale up. 
To this end, there has been increasing research interest to
investigate learning representations from unlabeled data. 
In particular, the image itself already contains structural information that can be utilized. 
One class of methods remove part of the visual data
and task the network with predicting what has been removed from the rest in a discriminative
manner~\cite{larsson2016learning,pathak2016context,trinh2019selfie,zhang2016colorful,zhang2017split}.
Another line of works treat each image (and augmentations of itself) as one class and use contrastive learning as self-supervision~\cite{bachman2019learning,bojanowski2017unsupervised,dosovitskiy2014discriminative,henaff2019data,hjelm2018learning,oord2018representation,wu2018unsupervised,he2019momentum,misra2019pirl,chen2020simple}.
Other self-supervised tasks include predicting rotation~\cite{gidaris2018unsupervised},
relative patch location~\cite{doersch2015unsupervised},
clusters~\cite{caron2018deep,caron2019leveraging}, and
number of objects~\cite{noroozi2017representation}, \etc.

\jcsu{
On top of those SSL tasks, combining different tasks can be beneficial~\cite{doersch2017multi}, in this work we also see its benefit.
Asano~\etal~\cite{asano2019surprising} showed that the representations can be learned with only one image and extreme augmentations. We also investigate SSL on a low-data regime, but use SSL as a regularizer instead of a pre-training task.
Goyal~\etal~\cite{goyal2019scaling} and Kolesnikov~\etal~\cite{kolesnikov2019revisiting} compared various SSL tasks at scale and concluded that solving jigsaw puzzles and predicting image rotations are among the most effective, motivating the choice of tasks in our experiments. Note that these two works did not include a comparison with contrastive learning approaches.}

In addition to pre-training models, SSL can also be used to improve other tasks. For example, Zhai~\etal~\cite{zhai2019s} showed that self-supervision can be used to improve recognition in a semi-supervised setting and presented results on a partially labeled version of the ImageNet dataset. Carlucci~\etal~\cite{carlucci2019domain} used self-supervision to improve domain generalization. In this work we use SSL to improve few-shot learning where the goal is to generalize to novel classes.

However, the focus of most prior works on SSL is to
supplant traditional supervised representation learning with
unsupervised learning on large unlabeled datasets for downstream
tasks.
Crucially in almost all prior works, self-supervised representations
consistently lag behind fully-supervised ones trained on
the same dataset with the same
architecture~\cite{goyal2019scaling,kolesnikov2019revisiting}.
\emph{In contrast, our work focuses on an important counterexample:} 
self-supervision can in fact augment standard supervised training for few-shot transfer learning in the low training data regime \emph{without} relying on any external dataset.

The most related work is that of Gidaris~\etal~\cite{gidaris2019boosting} who also use self-supervision to improve few-shot learning. 
\jcsu{
Although the initial results are similar (Table~\ref{tab:comparison}), we further show these benefits on several datasets with harder recognition problems (fine-grained classification) and with deeper models (Section~\ref{sec:comparison}). 
}
Moreover, we present a novel analysis of the impact of the domain of unlabeled data (Section~\ref{sec:domain}). 
Finally, we propose a new and simple approach to automatically select similar-domain unlabeled data for self-supervision (Section~\ref{sec:pool}).

\subsubsection{Multi-task learning}
Our work is related to multi-task learning, a class of techniques
that train on multiple task objectives together to improve each one.
Previous works in the computer vision literature have shown moderate
benefits by combining tasks such as edge, normal, and 
saliency estimation for images, or part segmentation and
detection for humans~\cite{kokkinos2017ubernet,maninis2019attentive,ren2018cross}.
However, there is significant evidence that training on multiple tasks together often hurts performance on
individual tasks~\cite{kokkinos2017ubernet,maninis2019attentive}. Only certain task combinations appear to be mutually beneficial, and sometimes specific architectures are needed. 
Our key contribution here is showing that self-supervised tasks and few-shot learning are indeed mutually beneficial in this sense. 




\subsubsection{Domain selection} On supervised learning, Cui
\etal~\cite{Cui_2018_CVPR} used Earth Mover's distance to measure the
domain similarity and select the source domain for pre-training. Ngiam
\etal~\cite{ngiam2018domain} found more data for pre-training does not
always help and proposed to use importance weights to select
pre-training data. Task2vec~\cite{achille2019task2vec} generates a
task embedding given a probe network and the target dataset. Such
embeddings can help select a better pre-training model from a pool of
experts which yields better performance after fine-tuning. Unlike
these, we do not assume that the source domain is labeled and rely on self-supervised learning.
On self-supervised learning, Goyal \etal~\cite{goyal2019scaling} used two
pre-training and target datasets to show the importance of
source domain on large-scale self-supervised learning.
Unlike this work, we investigate the performance on few-shot learning across a
number of domains, as well as investigate methods for domain selection.
A concurrent work~\cite{Wallace2020Extending} also investigates the effect of domain shifts on SSL.

  

\section{Method}
We adopt the commonly used setup for few-shot learning where one is
provided with labeled training data for a set of \emph{base}
classes $\mathcal{D}_b$ and a much smaller training set (typically 1-5
examples per class) for \emph{novel} classes $\mathcal{D}_n$.
The goal of the few-shot learner is to learn representations on the base
classes that lead to good generalization on novel classes.
Although in theory the base classes are assumed to have a large number
of labeled examples, in practice this number can be quite small for
novel or fine-grained domains, \eg less than 5000 images for the birds
dataset~\cite{cub}, making it challenging to learn a generalizable
representation.

Our framework, as seen in Fig.~\ref{fig:overview}, combines
\emph{meta-learning} approaches for few-shot learning with
\emph{self-supervised learning}.
Denote a labeled training dataset ${\cal D}_s$ as $\{(x_i,
y_i)\}_{i=1}^n$ consisting of pairs of images $x_i \in {\cal X}$ and
labels $y_i \in {\cal Y}$.
A feed-forward convolutional network $f(x)$ 
maps the input to an embedding space which is then mapped to
the label space using a classifier $g$.
The overall mapping from the input to the label can be
written as $g \circ f (x): {\cal X}\rightarrow {\cal Y}$.
Learning consists of estimating functions $f$ and $g$ that minimize
an empirical loss $\ell$ over the training data along with
suitable regularization ${\cal R}$ over the functions $f$ and $g$. 
This can be written as:
\[
 {\cal L}_s := \sum_{(x_i,y_i) \in {\cal D}_s} \ell\big(g\circ f (x_i), y_i\big) + {\cal R}(f, g).
\]
A commonly used loss is the cross-entropy loss and a regularizer is
the $\ell_2$ norm of the parameters of the functions.
In a transfer learning setting $g$ is discarded and relearned on
training data for novel classes.

We also consider self-supervised losses ${\cal L}_{ss}$
based on labeled data $x \rightarrow (\hat{x},\hat{y}$) that can be
derived automatically without any human labeling.
Fig.~\ref{fig:overview} shows two examples: the \emph{jigsaw task}
rearranges the input image and uses the index of the permutation as
the target label, while the \emph{rotation task} 
uses the angle of the rotated image as the target label. 
A separate function $h$ is used to predict these
labels from the shared feature backbone $f$ with a self-supervised loss:
\[
 {\cal L}_{ss} := \sum_{x_i \in {\cal D}_{ss}} \ell\big(h\circ f (\hat{x}_i), \hat{y}_i\big).
\]
Our final loss combines the two: ${\cal L} := {\cal L}_s + {\cal L}_{ss}$
and thus the self-supervised losses act as a data-dependent
regularizer for representation learning.
The details of these losses are described in the next sections.

Note that the domain of images used for supervised ${\cal D}_s$ and
self-supervised  ${\cal D}_{ss}$ losses need not to be identical. 
In particular, we would like to use larger sets of
images for self-supervised learning from related domains.
The key questions we ask are:
(1) How effective is SSL when ${\cal D}_s = {\cal D}_{ss}$
especially when we have a small sample of $D_s$?
(2) How do the domain shifts between \ds and \dss affect
generalization performance?
and (3) How to select images from a large, generic pool to construct
an effective \dss given a target domain \ds?

\subsection{Supervised Losses (${\cal L}_s$)}
\label{sec:sl}

Most of our results are presented using a meta-learner based on prototypical
networks~\cite{snell2017prototypical} that perform episodic training
and testing over sampled datasets in stages called meta-training and
meta-testing.
During meta-training, we randomly sample $N$ classes from the base set 
$\mathcal{D}_b$, then we select a support set $\mathcal{S}_b$ with $K$
images per class and another query set $\mathcal{Q}_b$ with $M$ images
per class. 
We call this an $N$-way $K$-shot classification task. 
The embeddings are trained to predict the labels of the query set
$\mathcal{Q}_b$ conditioned on the support set $\mathcal{S}_b$ using a
nearest mean (prototype) classifier. The objective is to minimize the
prediction loss on the query set.
Once training is complete, given the novel dataset $\mathcal{D}_n$, class
prototypes are recomputed for classification and query examples
are classified based on the distances to the class prototypes.

Prototypical networks are related to distance-based learners such as
matching networks~\cite{vinyals2016matching} or metric-learning based
on label similarity~\cite{koch2015siamese}.
We also present few-shot classification results using
a gradient-based meta-learner called MAML~\cite{finn2017model}, and one trained with a
standard cross-entropy loss on all the base classes. 
We also present standard classification results where the test set
contains images from the same base categories in Appendix~A.3.





\subsection{Self-supervised Losses (${\cal L}_{ss}$)}
\label{sec:ssl}
We consider two losses motivated by a recent
large-scale comparison of the effectiveness of self-supervised
learning tasks~\cite{goyal2019scaling} described below:
\begin{itemize}
\item \emph{Jigsaw puzzle task loss.} Here the input image $x$ is
tiled into 3$\times$3 regions and permuted randomly to obtain an input
$\hat{x}$. The target label $\hat{y}$ is the index of the permutation. 
The index (one of 9!) is reduced to one of 35 following the procedure
outlined in~\cite{noroozi2016unsupervised}, which grouped the 
possible permutations based on the hamming distance to control
the difficulty of the task.

\item \emph{Rotation task loss.} We follow the method
  of~\cite{gidaris2018unsupervised} where the input image $x$ is
  rotated by an angle $\theta \in
  \{0\degree,90\degree,180\degree,270\degree\}$ to obtain $\hat{x}$
  and the target label $\hat{y}$ is the index of the angle.
\end{itemize}

\noindent
In both cases we use the cross-entropy loss between the target and prediction.

\subsection{Stochastic Sampling and Training}
When the images used for SSL and meta-learning are
identical, \ie, \ds~=~\dss, the same batch of images are used for
computing both losses ${\cal L}_s$ and ${\cal L}_{ss}$.
For experiments investigating the effect of domain shifts described in
Section~\ref{sec:domain} and \ref{sec:pool}, where SSL and
meta-learner are trained on different domains, \ie \ds$\neq$~\dss, 
a separate batch of size of 64 is used for computing ${\cal L}_{ss}$. 
After the two forward passes, one for the supervised task and one for
the self-supervised task, the two losses are combined and gradient updates are performed.
While other techniques
exist~\cite{chen2017gradnorm,kendall2018multi,sener2018multi}, simply
averaging the two losses performed well.




\section{Experiments}\label{sec:exp}
We first describe the datasets and experimental details. In Section~\ref{sec:fsl}, we present the results of using SSL to improve few-shot learning on various datasets.
In Section~\ref{sec:domain}, we show the effect of domain shift between labeled and unlabeled data for SSL.
Last, we propose a way to select images from a pool for SSL to further improve the performance of few-shot learning in Section~\ref{sec:pool}. 

\begin{table}[!t]
\renewcommand{\arraystretch}{1.2}
\renewcommand{\tabcolsep}{2.5pt}
\begin{center}
\includegraphics[width=\linewidth]{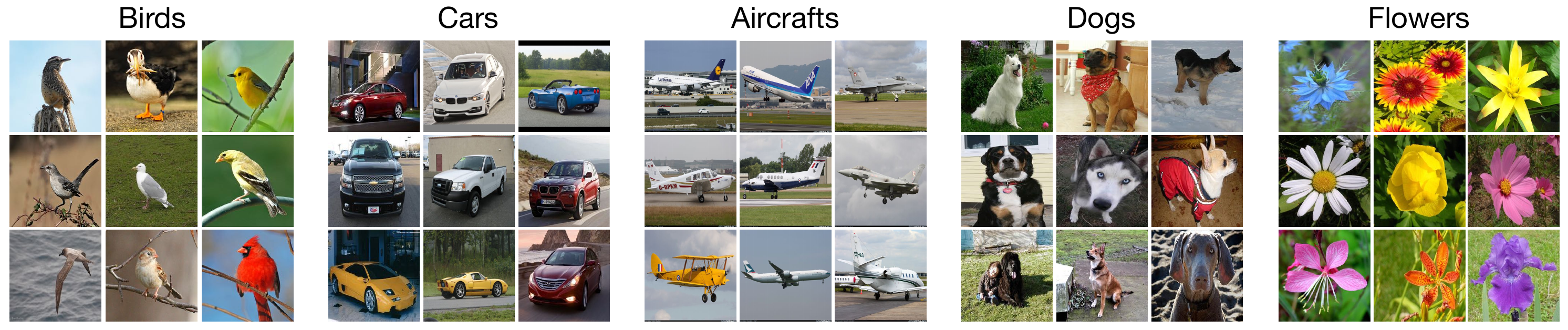}
   \begin{tabular}{c|c|c|c|c|c|c|c|c|c}
	\toprule
 		Setting & Set & Stats & \shortstack{\textit{mini}-\\ImageNet}  & \shortstack{\textit{tiered}-\\ImageNet} & Birds & Cars & Aircrafts & Dogs & Flowers \\
        \hline
		\multirow{6}{*}{\shortstack{Few-shot\\transfer}} & \multirow{2}{*}{Base} & classes & 64 & 351 & 100 & 98 & 50 & 60 & 51 \\
		 && images & 38,400 & 448,695 & 5885 & 8162 & 5000 & 10337 & 4129 \\
		 \cline{2-10}
		 &\multirow{2}{*}{Val} & classes & 16 & 97 & 50 & 49 & 25 & 30 & 26 \\
		 && images & 9,600 & 124,261 & 2950 & 3993 & 2500 & 5128 & 2113 \\
		 \cline{2-10}
		 &\multirow{2}{*}{Novel} & classes & 20 & 160 & 50 & 49 & 25 & 30 & 25  \\
		 && images & 12,000 & 206,209 & 2953 & 4030 & 2500 & 5115 & 1947 \\
		\bottomrule
	\end{tabular}
	\end{center}
\caption{\textbf{Example images and dataset statistics}. For few-shot learning experiments the classes are split into \emph{base}, \emph{val}, and \emph{novel} set. Image representations learned on \emph{base} set are evaluated on the \emph{novel} set while \emph{val} set is used for cross-validation.
These datasets vary in the number of classes but are orders of magnitude smaller than ImageNet dataset.}
\label{tab:stats}
\end{table}

\subsubsection{Datasets and benchmarks} We experiment with datasets across diverse
domains: Caltech-UCSD birds~\cite{cub}, Stanford cars~\cite{cars}, FGVC
aircrafts~\cite{aircrafts}, Stanford dogs~\cite{dogs},
and Oxford flowers~\cite{flowers}.
Each dataset contains between 100 and 200 classes with a few thousands of
images.
We also experiment with the widely-used
\textit{mini}-ImageNet~\cite{vinyals2016matching} and
\textit{tiered}-ImageNet~\cite{ren2018meta} benchmarks for few-shot
learning. In \textit{mini}-ImageNet, each class has 600 images, wherein \textit{tiered}-ImageNet each class has 732 to 1300 images.

We split classes within a dataset into three disjoint sets: \emph{base,}
\emph{val}, and \emph{novel}. 
For each class, all the images in the dataset are used in the
corresponding set.
A model is trained on the base set of categories, validated on the
val set,
and tested on the novel set of categories given a few examples per
class. 
For birds, we use the same split as~\cite{chen2019closerfewshot},
where \{\emph{base, val, novel}\} sets have \{100, 50, 50\} classes
respectively. 
The same ratio is used for the other four fine-grained datasets. 
We follow the original splits for
\textit{mini}-ImageNet and
\textit{tiered}-ImageNet. 
The statistics of various datasets used in our experiments are shown in Table~\ref{tab:stats}.
Notably, fine-grained datasets are significantly smaller.

We also present results on a setting where the base set is
``degraded'' either by (1) reducing the resolution, (2) removing
color, or (3) reducing the
number of training examples. 
This allows us to study the effectiveness of SSL on even smaller
datasets and as a function of the difficulty of the task.


\subsubsection{Meta-learners and feature backbone}
We follow the best practices and use the codebase for few-shot
learning described in \cite{chen2019closerfewshot}.
In particular, we use ProtoNet~\cite{snell2017prototypical} with a 
ResNet-18~\cite{he2016deep} network as the feature backbone.
Their experiments found this to be the best performing.
We also present experiments with other meta-learners such as MAML~\cite{finn2017model} and
softmax classifiers in Section~\ref{sec:fsl}.

\subsubsection{Learning and optimization}
We use 5-way (classes) and 5-shot (examples per-class) with 16 query
images for training. 
For experiments using 20\% of labeled data, we use 5 query images for training since the minimum number of images per class is 10. 
The models are trained with ADAM~\cite{kingma2014adam} with
a learning rate of 0.001 for 60,000 episodes. 
We report the mean accuracy and 95\% confidence interval over
600 test experiments. 
In each test episode, $N$ classes are selected from the novel set, and
for each class 5 support images and 16 query images are selected.
We report results for $N=\{5,20\}$ classes.


\subsubsection{Image sampling and data augmentation}
Data augmentation has a significant impact on few-shot
learning performance.
We follow the data augmentation procedure outlined
in~\cite{chen2019closerfewshot} which resulted in a strong baseline
performance.
For label and rotation predictions, images are first resized to 224
pixels for the shorter edge while maintaining the aspect ratio, from which a central
crop of 224$\times$224 is obtained. 
For jigsaw puzzles, we first randomly crop 255$\times$255 region from
the original image with random scaling between $[0.5,1.0]$, then split
into 3$\times$3 regions, from which a random crop of size 64$\times$64
is picked.
While it might appear that with self-supervision the model effectively sees more images, 
SSL provides consistent improvements even after extensive data augmentation
including cropping, flipping, and color jittering.
More experimental details are in Appendix~A.5.

\subsubsection{Other experimental results} In Appendix~A.3, we show the benefits of using self-supervision for \emph{standard} classification tasks when training the model \emph{from scratch}. We further visualize these models in Appendix~A.4 to show that models trained with self-supervision tend to avoid accidental correlation of background features to class labels.



\subsection{Results on Few-shot Learning}\label{sec:fsl}
\begin{figure}[t]
\centering
\includegraphics[width=\linewidth]{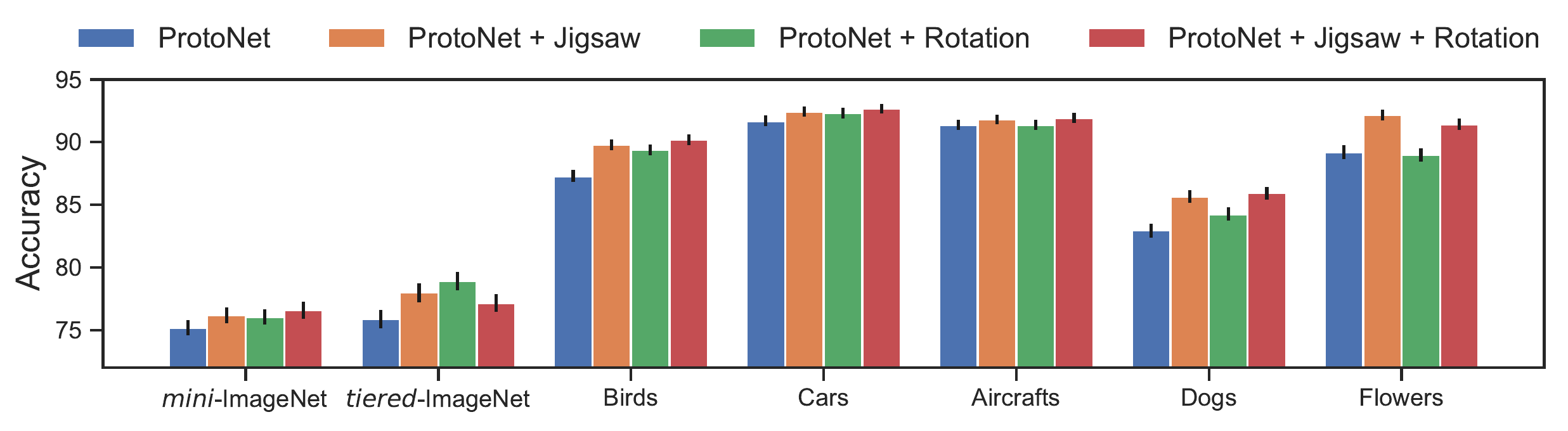}
\caption{\textbf{Benefits of SSL for few-shot learning tasks.} 
We show the accuracy of the ProtoNet baseline of using different SSL tasks.
  The jigsaw task results in an improvement of the
  5-way 5-shot classification accuracy across
  datasets. Combining SSL tasks can be beneficial for some datasets.
  Here SSL was performed on images within the base
  classes only.
  See Appendix~A.1 for a tabular version and results for 20-way 5-shot classification.}
\label{fig:fsl_result}
\end{figure}

\subsubsection{Self-supervised learning improves few-shot learning} 
Fig.~\ref{fig:fsl_result} shows the accuracies of various models on few-shot learning benchmarks.
Our ProtoNet baseline matches the results 
of the \textit{mini}-ImageNet and birds datasets presented
in~\cite{chen2019closerfewshot} (in their Table A5).
Our results show that jigsaw puzzle task improves the ProtoNet baseline on all
seven datasets. 
Specifically, it reduces the \emph{relative error rate} by
4.0\%, 8.7\%, 19.7\%, 8.4\%, 4.7\%, 15.9\%, and 27.8\% on
\textit{mini}-ImageNet, \textit{tiered}-ImageNet, birds, cars,
aircrafts, dogs, and flowers datasets respectively.
Predicting rotations also improves the ProtoNet baseline on
most of the datasets, except for aircrafts and flowers. We speculate this is because most flower images are symmetrical, and airplanes are usually horizontal, making the rotation task too hard or too trivial respectively to benefit the main task.
In addition, combining these two SSL tasks can be beneficial sometimes. 
A tabular version and the results of 20-way classification
are included in Appendix~A.1.

\begin{figure}[t]
\centering
\includegraphics[width=\linewidth]{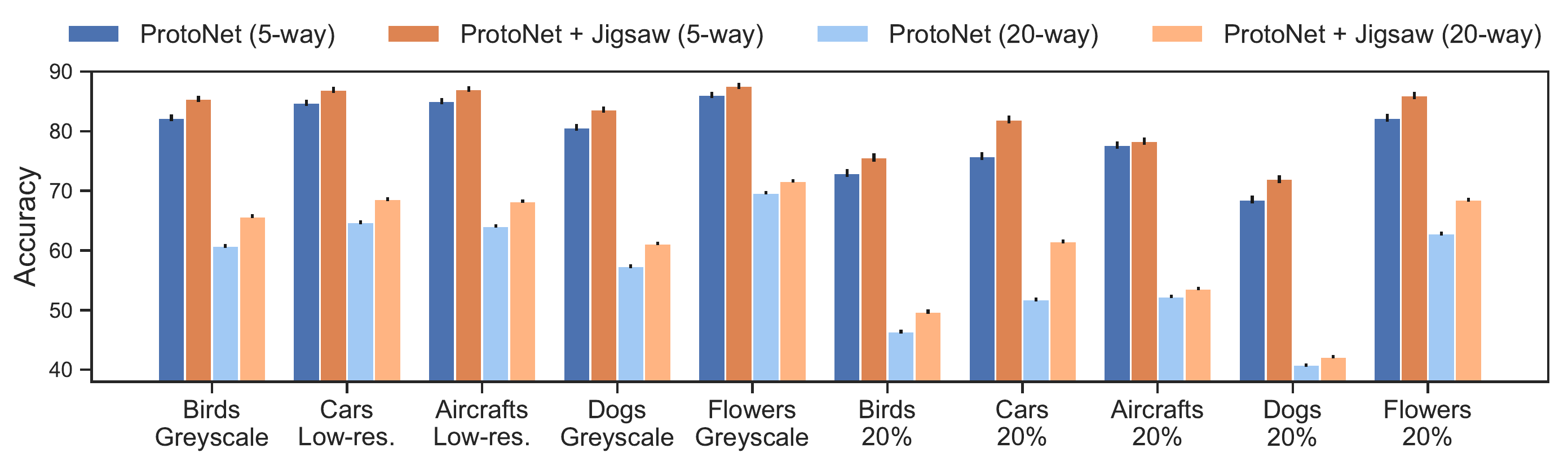}
\caption{\textbf{Benefits of SSL for
    \emph{harder} few-shot learning tasks.}
    We show the accuracy of using the jigsaw puzzle task over ProtoNet baseline on
    harder versions of the datasets. 
    We see that SSL is effective even on smaller datasets and the relative benefits are
    higher.
    \label{fig:low_quality_result}
    }
\end{figure}

\begin{table}[!t]
  \renewcommand{\arraystretch}{1.1}
  \setlength{\tabcolsep}{5.5pt}
  \begin{center}
  \begin{tabular}{c|c|c|c|c|c}
    \toprule
	\multirow{2}{*}{Loss}& Birds & Cars & Aircrafts & Dogs & Flowers\\
    \cline{2-6}
    &\multicolumn{5}{c}{5-way 5-shot} \\
    \hline
    Softmax & 81.5\rpm0.5&87.7\rpm0.5&89.2\rpm0.4&77.6\rpm0.6&91.0\rpm0.5\\
    Softmax + Jigsaw & {83.9\rpm0.5} & {90.6\rpm0.5} & {89.6\rpm0.4} & {77.8\rpm0.6} &91.1\rpm0.5\\
    MAML & 81.2\rpm0.7&86.9\rpm0.6&88.8\rpm0.5&77.3\rpm0.7&79.0\rpm0.9\\
    MAML + Jigsaw &  81.1\rpm0.7&{89.0\rpm0.5}&{89.1\rpm0.5}&77.3\rpm0.7&{82.6\rpm0.7}\\
    ProtoNet & 
    87.3\rpm0.5 &91.7\rpm0.4 &91.4\rpm0.4 &83.0\rpm0.6 &89.2\rpm0.6
    \\
    ProtoNet + Jigsaw & 
    \textbf{89.8\rpm0.4} & \textbf{92.4\rpm0.4} & \textbf{91.8\rpm0.4} & \textbf{85.7\rpm0.5} & \textbf{92.2\rpm0.4}\\
    \bottomrule
  \end{tabular}
  \end{center}
  \caption{\textbf{Performance on few-shot learning using different
      meta-learners.} Using jigsaw puzzle loss improves
    different meta-learners on most of the datasets. ProtoNet
    with jigsaw loss performs the best on all five datasets.}
  
  \label{tab:more_fsl_result}
\end{table}

\subsubsection{Gains are larger for harder tasks}
Fig.~\ref{fig:low_quality_result} shows the performance on the
degraded version of the same datasets (first five groups).
For cars and aircrafts we use low-resolution images where the images
are down-sampled by a factor of \emph{four} and up-sampled back to 224$\times$224 with
bilinear interpolation.
For natural categories we discard color.
Low-resolution images are considerably harder to classify for man-made
categories while color information is most useful for natural
categories~\cite{su2017adapting}.
On birds and dogs datasets, the improvements using
self-supervision (3.2\% and 2.9\% on 5-way 5-shot) are higher compared to color images (2.5\% and 2.7\%), similarly on the
cars and aircrafts datasets with low-resolution images (2.2\% and 2.1\% vs.~0.7\% and 0.4\%).
We also conduct an experiment where only 20\% of the images in the
base categories are used for both SSL and meta-learning (last five groups in Fig.~\ref{fig:low_quality_result}).
This results in a much smaller training set than standard few-shot benchmarks: 20\% of the birds dataset amounts to only roughly 3\% of the popular \textit{mini}-ImageNet dataset.
We find larger benefits from SSL in this setting.
For example, the gain from the jigsaw puzzle loss for 5-way 5-shot car classification increases from 0.7\% (original dataset) to 7.0\% (20\% training data).

\subsubsection{Improvements generalize to other meta-learners}
We combine SSL with other meta-learners and find the combination
to be effective.
In particular, we use MAML~\cite{finn2017model}
and a standard feature extractor trained with cross-entropy loss (softmax) as in~\cite{chen2019closerfewshot}.
Table~\ref{tab:more_fsl_result} compares meta-learners based on a ResNet-18 network trained with and without \emph{jigsaw puzzle loss}. 
We observe that the average 5-way 5-shot accuracies across five fine-grained datasets for softmax, MAML, and ProtoNet improve from 85.5\%, 82.6\%, and 88.5\% to 86.6\%, 83.8\%, and 90.4\% respectively when combined with the  jigsaw puzzle task.
Self-supervision improves performance across different meta-learners and different datasets; however,
ProtoNet trained with self-supervision is the best model across all datasets.

\begin{table}[t!]
  \setlength{\tabcolsep}{5pt}
    \begin{center}
    \begin{tabular}{c|c|c|c|c}
      \toprule
      \textbf{Model} & \textbf{mage Size}  & \textbf{Backbone} & \textbf{SSL} & \textbf{Accuracy (\%)}\\
    \hline
      MAML~\cite{finn2017model} & \multirow{8}{*}{84$\times$84} & Conv4-64 & - & 63.1 \\
      ProtoNet~\cite{snell2017prototypical} & & Conv4-64 & - & 68.2 \\
      RelationNet~\cite{Sung_2018_CVPR} & & Conv4-64 & - & 65.3 \\
      LwoF~\cite{gidaris2018dynamic} & & Conv4-64 & - & 72.8 \\
      PFA~\cite{qiao2018few}$^*$ & & WRN-28-10 & - & 73.7 \\
      TADAM~\cite{oreshkin2018tadam} & & ResNet-12 & - & 76.7 \\
      LEO~\cite{rusu2018meta}$^*$ & & WRN-28-10 & - & 77.6 \\
      MetaOptNet-SVM~\cite{lee2019meta}$^\dagger$ & & ResNet-12 & - & 78.6\\
    \hline
      \multirow{2}{*}{\shortstack{Chen \etal~\cite{chen2019closerfewshot}\\(ProtoNet)}}
      & 84$\times$84 &  Conv4-64 & - & 64.2 \\
      & 224$\times$224 &  ResNet-18 & - & 73.7 \\
    \hline
      \multirow{6}{*}{\shortstack{Gidaris \etal~\cite{gidaris2019boosting}\\
      (ProtoNet)}} &\multirow{6}{*}{84$\times$84}&\multirow{2}{*}{Conv4-64}&-&70.0\\
      &&&Rotation&71.7\\
      \cline{3-5}
      &&\multirow{2}{*}{Conv4-512}&-&71.6\\
      &&&Rotation&74.0\\
      \cline{3-5}
      &&\multirow{2}{*}{WRN-28-10}&-&68.7\\
      &&&Rotation&72.1\\
    \hline
      \multirow{4}{*}{\shortstack{\textbf{Ours}\\(ProtoNet)}}&\multirow{4}{*}{224$\times$224}&\multirow{4}{*}{ResNet-18}&-&75.2\\
      &&&Rotation&76.0\\
      &&&Jigsaw&76.2\\
      &&&Rot.+Jig.&76.6\\
    \bottomrule
    \end{tabular}
    \end{center}
    \caption{\textbf{Comparison with prior works on \textit{mini}-ImageNet.} 5-shot 5-way classification accuracies on 600 test episodes are reported. The implementation details including image size, backbone model, and training are different in each paper. $^*$validation classes are used for training. $^\dagger$dropblock~\cite{ghiasi2018dropblock}, label smoothing, and weight decay are used.}
    \label{tab:comparison}
\end{table}

\subsubsection{Self-supervision alone is not enough}
SSL alone significantly lags behind supervised learning in our
experiments. 
For example, a ResNet-18 trained with SSL alone achieve 32.9\% (w/ jigsaw) and 33.7\% (w/ rotation) 5-way 5-shot accuracy averaged
across five fine-grained datasets. While this is better than a random initialization
(29.5\%), it is dramatically worse than one trained with a simple
cross-entropy loss (85.5\%) on the labels (details
in Table~\ref{tab:fsl_result} in Appendix~A.1).
Surprisingly, we also found that initialization with SSL
followed by meta-learning did \textit{not} yield improvements over
meta-learning starting from random initialization, supporting the view that SSL acts as a feature
regularizer.

\subsubsection{Few-shot learning as an evaluation for
    self-supervised tasks} The few-shot classification task provides a
way of evaluating the effectiveness of self-supervised tasks. For example, on 5-way 5-shot aircrafts classification, training
with only jigsaw and rotation task gives 38.8\% and 29.5\% respectively,
suggesting that rotation is not an effective self-supervised task for
airplanes.
We speculate that it might be because the task is too easy as
airplanes are usually horizontal.


\subsubsection{Comparison with prior works}\label{sec:comparison}
Our results also echo those of~\cite{gidaris2019boosting} who find that the rotation task improves on \textit{mini}- and \textit{tiered}-ImageNet.
In addition we show the improvement still holds when using deeper networks, higher resolution images, and in fine-grained domains. We provide a comparison with other few-shot learning methods in Table~\ref{tab:comparison}.

\subsection{Analyzing the Effect of Domain Shift for Self-supervision}
\label{sec:domain}

Scaling SSL to massive unlabeled datasets that are readily available for some domains is a promising avenue for improvement.
\emph{However, do more unlabeled data always help for a task in hand?}
This question hasn't been sufficiently addressed in the literature as
most prior works study the effectiveness of SSL on a curated set of
images, such as ImageNet, and their transferability to a handful of tasks.
We conduct a series of experiments to characterize
the effect of size and distribution ${\cal D}_{ss}$ of images used for SSL in the context of few-shot learning on domain ${\cal D}_s$. 

\begin{figure}[t!]
  \captionsetup[subfigure]{}
  \newcommand\wi{.49\linewidth}
  \newcommand\wifig{\linewidth}
  \centering
  \begin{subfigure}{\wi}
    \centering 
    \includegraphics[clip, width=\wifig]{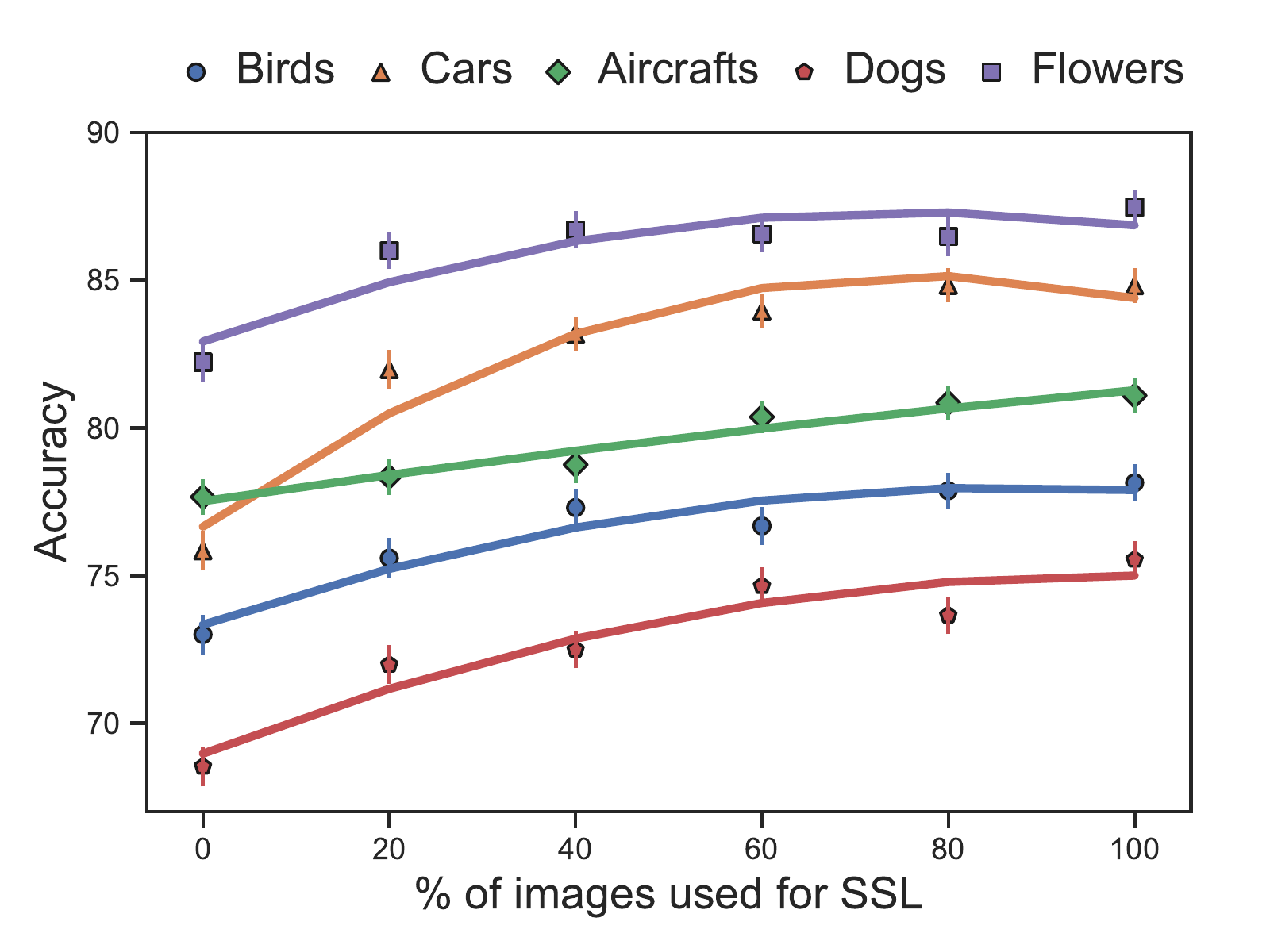}
    \caption{Effect of number of images on SSL.}
    \label{fig:vary}
  \end{subfigure}
  \begin{subfigure}{\wi}
    \centering
    \includegraphics[clip, width=\wifig]{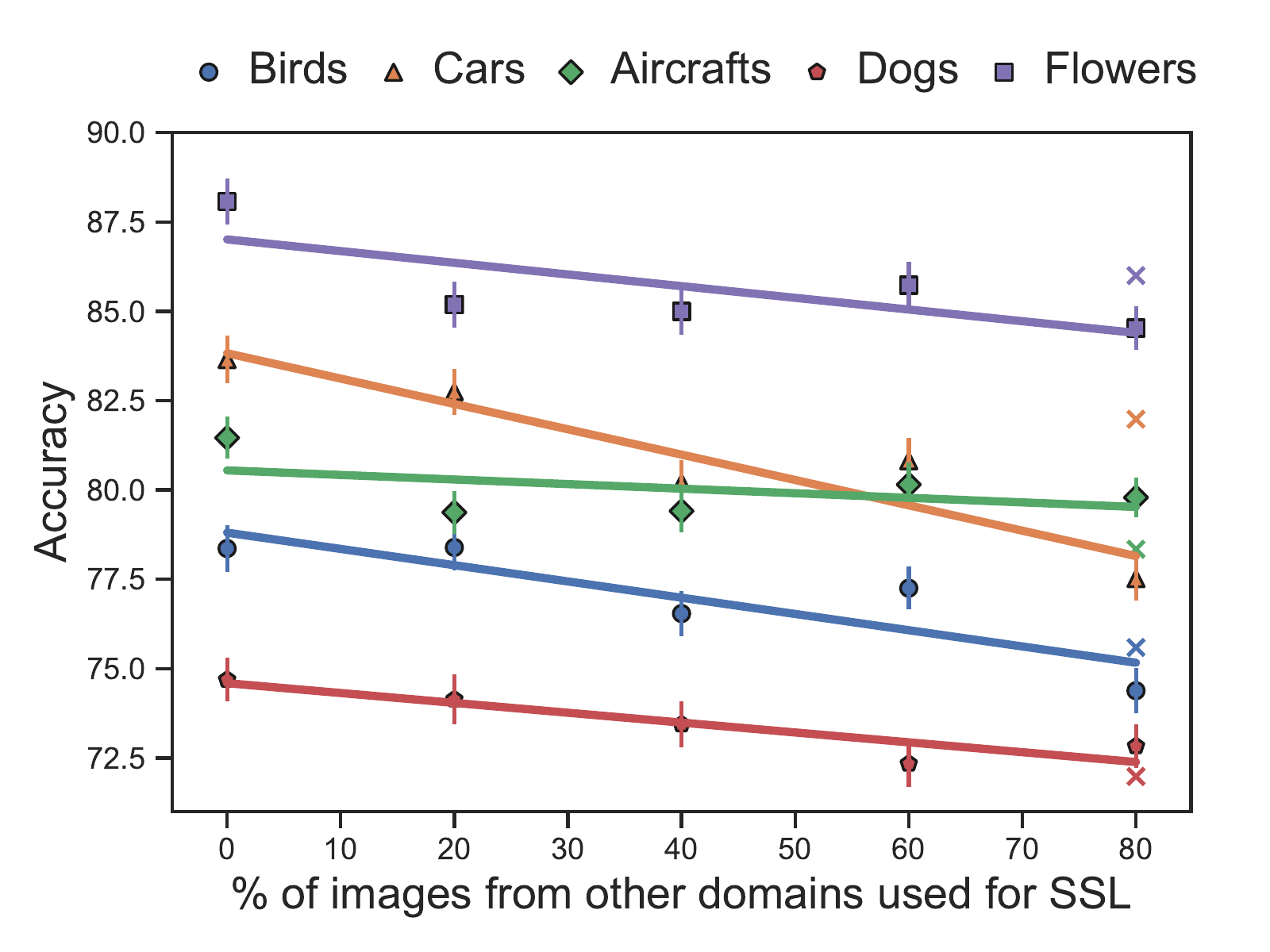}
    \caption{Effect of domain shift on SSL.}
    \label{fig:mixed}
  \end{subfigure}
  \caption{\textbf{Effect of size and domain of SSL on 5-way 5-shot classification accuracy.}
    \textbf{(a)}
    More unlabeled data from the same domain for SSL improves
    the performance of the meta-learner. 
    \textbf{(b)}
    Replacing a fraction (x-axis) of the images with
    those from other domains makes SSL less effective.
  }
  \label{fig:domain_all}
\end{figure}

First, we investigate if SSL on unlabeled data from the same domain 
improves the meta-learner. 
We use 20\% of the images in the base categories for
meta-learning identical to the setting in Fig.~\ref{fig:low_quality_result}. 
The labels of the remaining 80\% data are withheld and only the images are used
for SSL. 
We systematically vary the number of images used by SSL from 20\%
to 100\%. 
The results are presented in Fig.~\ref{fig:vary}.
The accuracy improves with the size of the unlabeled set with diminishing returns.
Note that 0\% corresponds to no SSL and 20\% corresponds to using only the 
labeled images for SSL (${\cal D}_{s} = {\cal D}_{ss}$).

Fig.~\ref{fig:mixed} shows an experiment where a fraction of the
unlabeled images are replaced with images from other four datasets.
For example, 20\% along the x-axis for birds indicate that 20\% of the images in the base set are replaced by images drawn uniformly at random from other datasets.
Since the numbers of images used for SSL is identical, the x-axis from left to right
represents increasing amounts of domain shifts between ${\cal D}_s$
and ${\cal D}_{ss}$.
We observe that the effectiveness of SSL decreases as the fraction of out-of-domain images
increases.
Importantly, training with SSL on the available 20\% within domain images (shown as crosses) is often (on 3 out of 5 datasets) better than increasing the set of images by five times to include out of domain images.


\subsection{Selecting Images for Self-supervision}\label{sec:pool}

\begin{figure}[t!]
\newcommand\wifig{\linewidth}
    \begin{center} 
    \includegraphics[clip, width=\wifig]{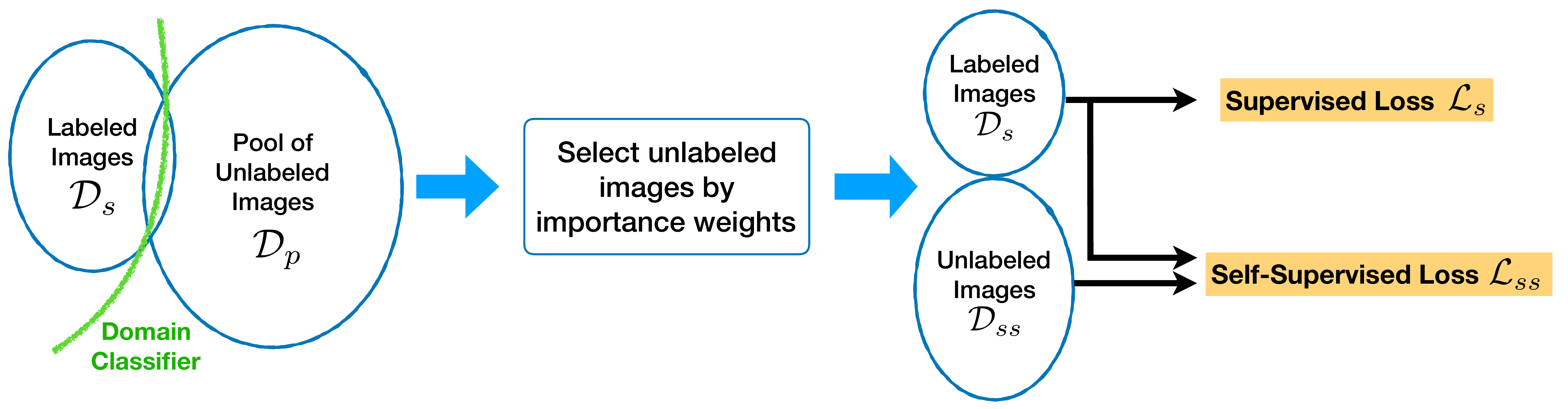}
    \includegraphics[clip, width=\wifig]{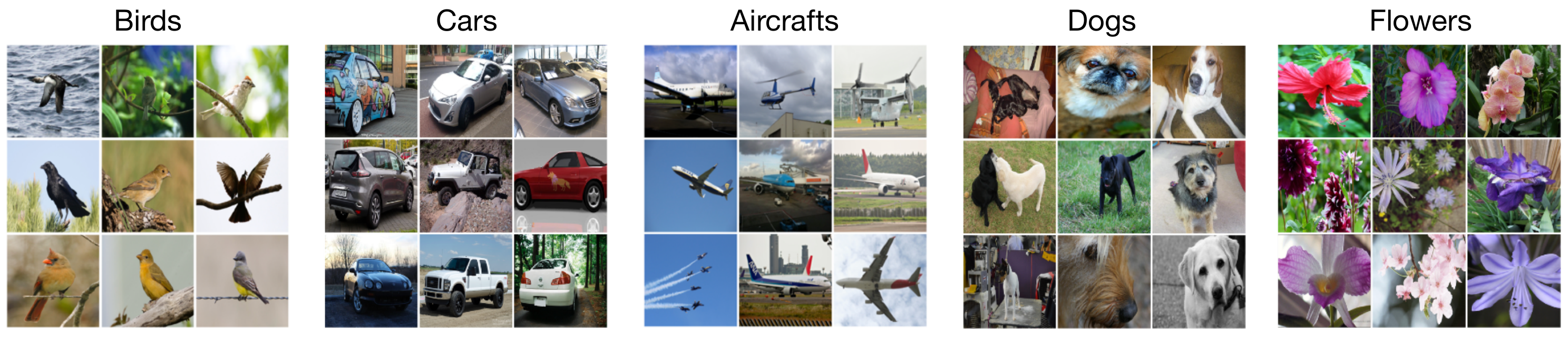}
    \end{center} 
    \caption{\textbf{Overview of domain selection for self-supervision.} \textbf{Top:} We first train a domain classifier using ${\cal D}_s$ and (a subset of) ${\cal D}_p$, then select images using the predictions from the domain classifier for self-supervision. \textbf{Bottom:} Selected images of each dataset using importance weights.}
    \label{fig:selected}
\end{figure}

\begin{figure}[t!]
\newcommand\wifig{\linewidth}
    \begin{center} 
    \includegraphics[clip, width=\wifig]{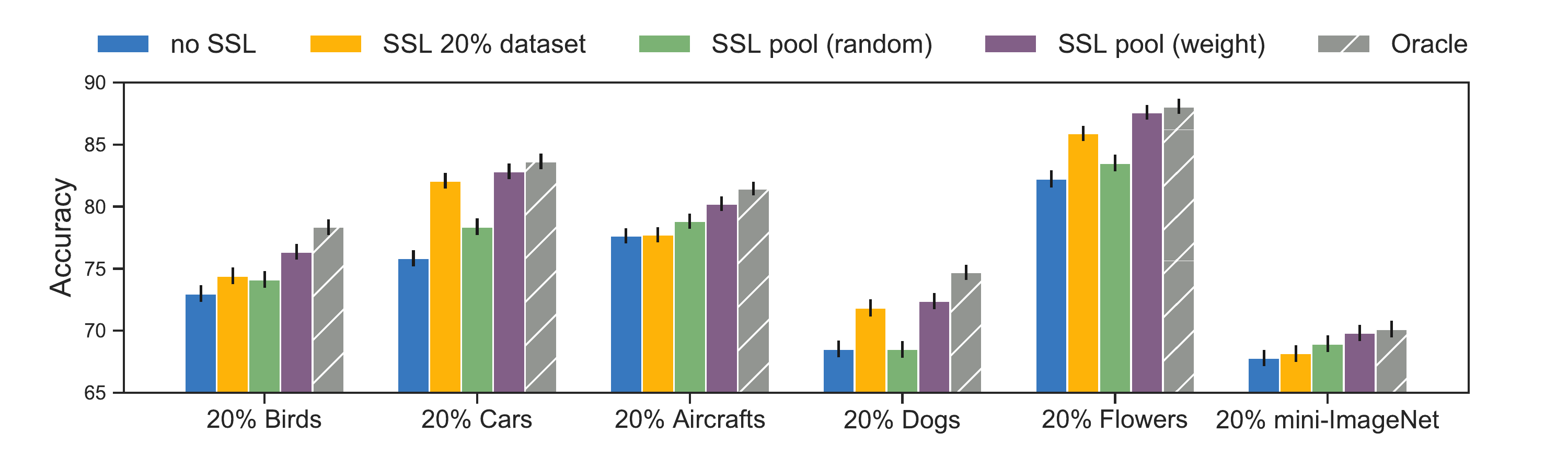}
    \end{center} 
    \caption{\textbf{Effectiveness of selected images for SSL.} With random selection, the extra unlabeled data often hurts the performance, while those sampled using the \emph{importance weights} improve performance on all five datasets. A tabular version is shown in Appendix~A.2.}
    \label{fig:pool}
\end{figure}

Based on the above analysis we propose a simple method to select images for SSL from a large, generic pool of unlabeled images in a dataset dependent manner.
We use a ``domain weighted'' model to select the top images based on a domain classifier, in our case a binary logistic regression model trained with images from the source domain ${\cal D}_s$ as the positive class and images from the pool ${\cal D}_{p}$ as the negative class based on ResNet-101 image features. 
The top images are selected according to the ratio $p(x \in {\cal D}_s)/p(x \in {\cal D}_{p})$. 
Note that these \emph{importance weights} account for the domain shift.
Fig.~\ref{fig:selected} shows an overview of the selection process.

We evaluate this approach using a pool of images ${\cal D}_{p}$ consisting of (1) the training images of the ``bounding box" subset of Open Images V5~\cite{OpenImages} which has 1,743,042 images from 600 classes, and (2) iNaturalist 2018 dataset~\cite{gvanhorn2018inat} which has 461,939 images from 8162 species. 
For each dataset, we use 20\% of the labeled images as ${\cal D}_s$.
The rest 80\% of the data are only used as the ``oracle'' where the unlabeled data are drawn from the exact same distribution as ${\cal D}_s$.
We show some of the selected images for self-supervision ${\cal D}_{ss}$ in Fig.~\ref{fig:selected}.


Fig.~\ref{fig:pool} shows the results of ProtoNet trained on 20\% labeled examples with jigsaw puzzle as self-supervision.
To have a fair comparison, for methods of selecting images from the pool, we select the same number (80\% of the original labeled dataset size) of images as ${\cal D}_{ss}$. 
We report the mean accuracy of five runs.
``SSL with 20\% dataset'' denotes a baseline of only using ${\cal D}_s$ for self-supervision (${\cal D}_{s}={\cal D}_{ss}$), which is our reference ``lower bound''.
SSL pool ``(random)'' and ``(weight)'' denote two approaches of selecting images for self-supervision.
The former selects images uniformly at random, which is detrimental for cars, dogs, and flowers. 
The pool selected according to the \emph{importance weights} provides significant improvements over ``no SSL'', ``SSL with 20\% dataset'', and ``random selection'' baselines on all five datasets.
The oracle is trained with the remaining 80\% of the original dataset as ${\cal D}_{ss}$, which is a reference ``upper bound''.

\section{Conclusion}
Self-supervision improves the performance on few-shot learning tasks across a range of different domains.
Surprisingly, we found that self-supervision is more beneficial for more challenging problems,
especially when the number of images used for self-supervision is small,
orders of magnitude smaller than previously reported results.
This has a practical benefit that the images within small datasets can
be used for self-supervision without relying on a large-scale external dataset. 
We have also shown that additional unlabeled images can improve performance only if they are from the \emph{same or similar} domains.
Finally, for domains where unlabeled data is limited, we present a novel, simple approach to automatically identify such similar-domain images from a larger pool.

Future work could investigate if using other self-supervised tasks can
also improve few-shot learning, in particular constrastive learning approaches~\cite{bachman2019learning,he2019momentum,henaff2019data,misra2019pirl,Tian2019}.
Future work could also investigate how and when self-supervision
improves generalization across self-supervised and supervised tasks empirically~\cite{achille2019task2vec,zamir2018taskonomy}.

\section*{Acknowledgement}
This project is supported in part by NSF \#1749833 and a DARPA LwLL grant. Our experiments were performed on the University of Massachusetts Amherst GPU cluster obtained under the Collaborative Fund managed by the Massachusetts Technology Collaborative.

\bibliographystyle{splncs04}
\bibliography{reference,reference_ssl}


\appendix
\section{Appendix}
In Appendix~\ref{appendix:more_fsl_result} and Appendix~\ref{appendix:pool_detail}, we provide all the numbers of the figures in Section~\ref{sec:fsl} and Section~\ref{sec:pool} separately. 
We show that SSL can also improve traditional fine-grained classification in Appendix~\ref{appendix:fgvc} and its model visualization in Appendix~\ref{appendix:visualization}. 
Last, we describe the implementation details in Appendix~\ref{appendix:exp_details}.

\subsection{Results on Few-shot Learning}\label{appendix:more_fsl_result}
\definecolor{mygray}{gray}{0.6}
\newcommand\grey[1]{\textcolor{mygray}{\textit{#1}}}
\begin{table}[!h]
  \renewcommand{\arraystretch}{1.2}
  \setlength{\tabcolsep}{0.9pt}
  \begin{center}
  \begin{tabular}{c|c|c|c|c|c|c|c}
    \toprule
    \multirow{2}{*}{Loss}& \shortstack{\textit{mini}-\\ImageNet} & \shortstack{\textit{tiered}-\\ImageNet} & Birds & Cars & Aircrafts & Dogs & Flowers\\
    \cline{2-8}
    &\multicolumn{7}{c}{5-way 5-shot} \\
    \hline
    ProtoNet(PN) & 
    75.2\rpm0.6 & 75.9\rpm0.7 & 87.3\rpm0.5 &91.7\rpm0.4 &91.4\rpm0.4 &83.0\rpm0.6 &89.2\rpm0.6
    \\
    PN+Jigsaw & {76.2\rpm0.6} & 78.0\rpm0.7 & {89.8\rpm0.4}&{92.4\rpm0.4}&{91.8\rpm0.4}&{85.7\rpm0.5}&\textbf{92.2\rpm0.4}
    \\
    \grey{Rel.~err.~red.} &\grey{4.0\%} &\grey{8.7\%}& \grey{19.7\%}& \grey{8.4\%} &\grey{4.7\%}& \grey{15.9\%}& \grey{27.8\%}\\
    Jigsaw & 25.6\rpm0.5 & 24.9\rpm0.4& 25.7\rpm0.5&25.3\rpm0.5&38.8\rpm0.6&24.3\rpm0.5&50.5\rpm0.7%
    \\
    PN+Rotation & 76.0\rpm0.6 & \textbf{78.9\rpm0.7} &  89.4\rpm0.4&92.3\rpm0.4&91.4\rpm0.4&84.3\rpm0.5&89.0\rpm0.5
    \\
    Rotation & 51.4\rpm0.7 & 50.7\rpm0.8&  33.1\rpm0.6&29.4\rpm0.5&29.5\rpm0.5&27.3\rpm0.5&49.4\rpm0.7%
    \\
    PN+Jig.+Rot. & \textbf{76.6\rpm0.7} & 77.2\rpm0.7 & \textbf{90.2\rpm0.4} & \textbf{92.7\rpm0.4} & \textbf{91.9\rpm0.4} & \textbf{85.9\rpm0.5} & 91.4\rpm0.5
    \\
    None & 31.0\rpm0.5 & 28.9\rpm0.5&  26.7\rpm0.5&25.2\rpm0.5&28.1\rpm0.5&25.3\rpm0.5&42.3\rpm0.8%
    \\
    \hline
    &\multicolumn{7}{c}{20-way 5-shot}\\
    \hline
    ProtoNet(PN) & 46.6\rpm0.3 & 49.7\rpm0.4 & 69.3\rpm0.3&78.7\rpm0.3&78.6\rpm0.3&61.6\rpm0.3&75.4\rpm0.3%
    \\
    PN+Jigsaw & 47.8\rpm0.3 & \textbf{52.4\rpm0.4} &  
    {73.7\rpm0.3}&79.1\rpm0.3&\textbf{79.1\rpm0.2}&{65.4\rpm0.3}&\textbf{79.2\rpm0.3}%
    \\
    Jigsaw & 9.2\rpm0.2& 7.5\rpm0.1&  8.1\rpm0.1&7.1\rpm0.1&15.4\rpm0.2&7.1\rpm0.1&25.7\rpm0.2%
    \\
    PN+Rotation & {48.2\rpm0.3} & \textbf{52.4\rpm0.4} & 
    72.9\rpm0.3&\textbf{80.0\rpm0.3}&78.4\rpm0.2&63.4\rpm0.3&73.9\rpm0.3
    \\
    Rotation & 27.4\rpm0.2 & 25.7\rpm0.3&  12.9\rpm0.2&9.3\rpm0.2&9.8\rpm0.2&8.8\rpm0.1&26.3\rpm0.2%
    \\
    PN+Jig.+Rot. & \textbf{49.0\rpm0.3} & 51.2\rpm0.4 & \textbf{75.0\rpm0.3} & 79.8\rpm0.3 & 79.0\rpm0.2 & \textbf{66.2\rpm0.3} & 78.6\rpm0.3
    \\
    None & 10.8\rpm0.1 & 11.0\rpm0.2&  9.3\rpm0.2&7.5\rpm0.1&8.9\rpm0.1&7.8\rpm0.1&22.6\rpm0.2%
    \\
    \bottomrule
  \end{tabular}
  \end{center}
  \caption{\textbf{Performance on few-shot learning tasks.} The mean accuracy (\%) and the 95\% confidence interval of 600 randomly chosen test experiments are reported for various combinations of loss functions. 
    The top part shows the accuracy on 5-way 5-shot classification tasks, while the bottom part shows the same on 20-way 5-shot. 
    Adding self-supervised losses to the ProtoNet loss improves the performance on all seven datasets on 5-way classification results. 
    On 20-way classification, the improvements are even larger.
    The last row indicates results with a randomly initialized network. The top part of this table corresponds to Figure~2 in Section~\ref{sec:fsl}.}
  \label{tab:fsl_result}
\end{table}

\begin{table}[!t]
  \renewcommand{\arraystretch}{1.2}
  \setlength{\tabcolsep}{1.3pt}
  \begin{center}
  \begin{tabular}{c|c|c|c|c|c}
    \toprule
    \multirow{2}{*}{Loss} & Birds & Cars & Aircrafts & Dogs & Flowers\\
    & Greyscale & Low-resolution & Low-resolution & Greyscale & Greyscale \\
    \hline
    &\multicolumn{5}{c}{5-way 5-shot}\\
    \hline
    ProtoNet & 82.2\rpm0.6 & 84.8\rpm0.5 & 85.0\rpm0.5 & 80.7\rpm0.6 & 86.1\rpm0.6\\
    ProtoNet + Jigsaw & 85.4\rpm0.6& 87.0\rpm 0.5 & 87.1\rpm0.5
    & 83.6\rpm0.5 & 87.6\rpm0.5\\
    \hline
    &\multicolumn{5}{c}{20-way 5-shot}\\
    \hline
    ProtoNet & 60.8\rpm0.4 & 64.7\rpm0.3 & 64.1\rpm0.3 & 57.4\rpm0.3 & 69.7\rpm0.3\\
    ProtoNet + Jigsaw & 65.7\rpm0.3 & 68.6\rpm0.3 & 68.3\rpm0.3 & 61.2\rpm0.3 & 71.6\rpm0.3\\
    \bottomrule
    \toprule
    Loss&20\% Birds&20\% Cars&20\% Aircrafts&20\% Dogs&20\% Flowers\\
    \hline
    &\multicolumn{5}{c}{5-way 5-shot}\\
    \hline
    ProtoNet & 73.0\rpm0.7 & 75.8\rpm0.7 & 77.7\rpm0.6 & 68.5\rpm0.7 & 82.2\rpm0.7\\
    ProtoNet + Jigsaw & 75.4\rpm0.7&82.8\rpm0.6&78.4\rpm0.6&69.1\rpm0.7&86.0\rpm0.6\\
    \hline
    &\multicolumn{5}{c}{20-way 5-shot}\\
    \hline
    ProtoNet & 46.4\rpm0.3&51.8\rpm0.4&52.3\rpm0.3&40.8\rpm0.3&62.8\rpm0.3
\\
    ProtoNet + Jigsaw &49.8\rpm0.3&61.5\rpm0.4&53.6\rpm0.3&42.2\rpm0.3&68.5\rpm0.3\\
    \bottomrule
  \end{tabular}
  \end{center}
  \caption{\textbf{Performance on \textit{harder} few-shot learning tasks.}
    Accuracies are reported on novel set for 5-way 5-shot and 20-way
    5-shot classification
    with degraded inputs, and with a subset (20\%) of the images in the base set. 
    The loss of color or resolution, and the smaller training set size make the tasks more challenging as seen by the drop in the performance of the ProtoNet baseline. However the improvements of using the \emph{jigsaw puzzle loss} are higher in comparison to the results presented in Table~\ref{tab:fsl_result}.
    \label{tab:low_quality_result}}
\end{table}

Table~\ref{tab:fsl_result} shows the performance of ProtoNet with different self-supervision on seven datasets. 
We also test the accuracy of the model on novel classes when trained \emph{only} with
self-supervision on the base set of images. 
Compared to the randomly initialized model (``None'' rows), training the network to predict rotations
gives around 2\% to 21\% improvements on all datasets, while
solving jigsaw puzzles only improves on aircrafts and flowers. 
However, these numbers are significantly worse than learning with
supervised labels on the base set, in line with the current
literature.

Table~\ref{tab:low_quality_result} shows the performance of ProtoNet with \emph{jigsaw puzzle loss} on harder benchmarks. The results on the degraded version of the datasets are shown in the top part, and the bottom part shows the results of using only 20\% of the images in the base categories. 
The gains using SSL are higher in this setting.

\subsection{Results on Selecting Images for SSL}\label{appendix:pool_detail}
Table~\ref{tab:selected} shows the performance of selecting images for self-supervision, a tabular version of Figure~\ref{fig:selected} in Section~\ref{sec:pool}.
``Pool (random)" uniformly samples images proportional to the size of each dataset, while the ``pool (weight)" one tends to pick more images from related domains.

\begin{table}[t!]
  \definecolor{Gray}{rgb}{0.95,0.95,0.95}
  \renewcommand{\arraystretch}{1.2}
    \begin{center}
    \begin{tabular}{c|c|c|c|c|c|c}
      \toprule
      \multirow{2}{*}{Method} & 20\% & 20\% & 20\% & 20\% & 20\% & 20\%~\textit{mini}-
      \\
      & Birds & Cars & Aircrafts & Dogs & Flowers & ImageNet\\
      \hline
    No SSL & 73.0\rpm0.7&75.8\rpm0.7&77.7\rpm0.6&68.5\rpm0.7&82.2\rpm0.7&67.81\rpm0.65\\
	SSL 20\% dataset &74.4\rpm0.7&82.1\rpm0.6&77.7\rpm0.6&71.8\rpm0.7&85.9\rpm0.6&68.47\rpm0.66\\
	SSL Pool (random) &74.1\rpm0.7&78.4\rpm0.7&78.8\rpm0.6&68.5\rpm0.7&83.5\rpm0.7&68.94\rpm0.68\\
	SSL Pool (weight) &\textbf{76.4\rpm0.6}&\textbf{82.9\rpm0.6}&\textbf{80.2\rpm0.6}&\textbf{72.4\rpm0.7}&\textbf{87.6\rpm0.6}&\textbf{69.81\rpm0.65}\\
	\grey{SSL 100\% (oracle)} &\grey{78.4\rpm0.6}&\grey{83.7\rpm0.6}&\grey{81.5\rpm0.6}&\grey{74.7\rpm0.6}&\grey{88.1\rpm0.6}&\grey{70.13\rpm0.67}\\
    \bottomrule
    \end{tabular}
    \end{center}
    \caption{\textbf{Performance on selecting images for self-supervision.} Adding more unlabeled images selected randomly from a pool often hurts the performance. Selecting similar images by importance weights improves on all five datasets.}
    \label{tab:selected}
\end{table}

\subsection{Results on Standard Fine-grained Classification}\label{appendix:fgvc}
Here we present results on standard fine-grained classification tasks. 
Different from few-shot transfer learning, all the classes are seen in the training set and the test set contains novel images from the same classes. 
We use the standard training and test splits provided in the datasets.
We investigate if SSL can improve the training of deep networks (\eg ResNet-18 network) when \emph{trained
from scratch} (\ie with random initialization) using images and labels in the training set only. 
The accuracy of using various loss functions are shown in Table~\ref{tab:fgvc}.
Training with self-supervision improves performance across datasets. 
On birds, cars, and dogs, predicting rotation gives 4.1\%, 3.1\%, and
3.0\% improvements, while on aircrafts and flowers, the \emph{jigsaw puzzle loss} yields 0.9\% and 3.6\% improvements.
\begin{table}[t!]
  \renewcommand{\arraystretch}{1.2}
  \setlength{\tabcolsep}{8pt}
    \begin{center}
    \begin{tabular}{c|c|c|c|c|c}
      \toprule
      Loss & Birds & Cars & Aircrafts & Dogs & Flowers\\
      \hline
      Softmax & 47.0 & 72.6 & 69.9 & 51.4 & 72.8\\
      Softmax + Jigsaw~~~ & 49.2 & 73.2 & \textbf{70.8} & 53.5 & \textbf{76.4}\\
      Softmax + Rotation & \textbf{51.1} & \textbf{75.7} & 70.0 & \textbf{54.4} & 73.5\\
      \bottomrule
    \end{tabular}
    \end{center}
    \caption{\textbf{Performance on standard fine-grained classification tasks.} Per-image accuracy (\%) on the test set are reported. Using self-supervision improves the accuracy of a ResNet-18 network trained \emph{from scratch} over the baseline of supervised training with cross-entropy (softmax) loss on all five datasets.}
    \label{tab:fgvc}
\end{table}

\subsection{Visualization of Learned Models}\label{appendix:visualization}
\begin{figure}[t]
\centering
\includegraphics[width=\linewidth]{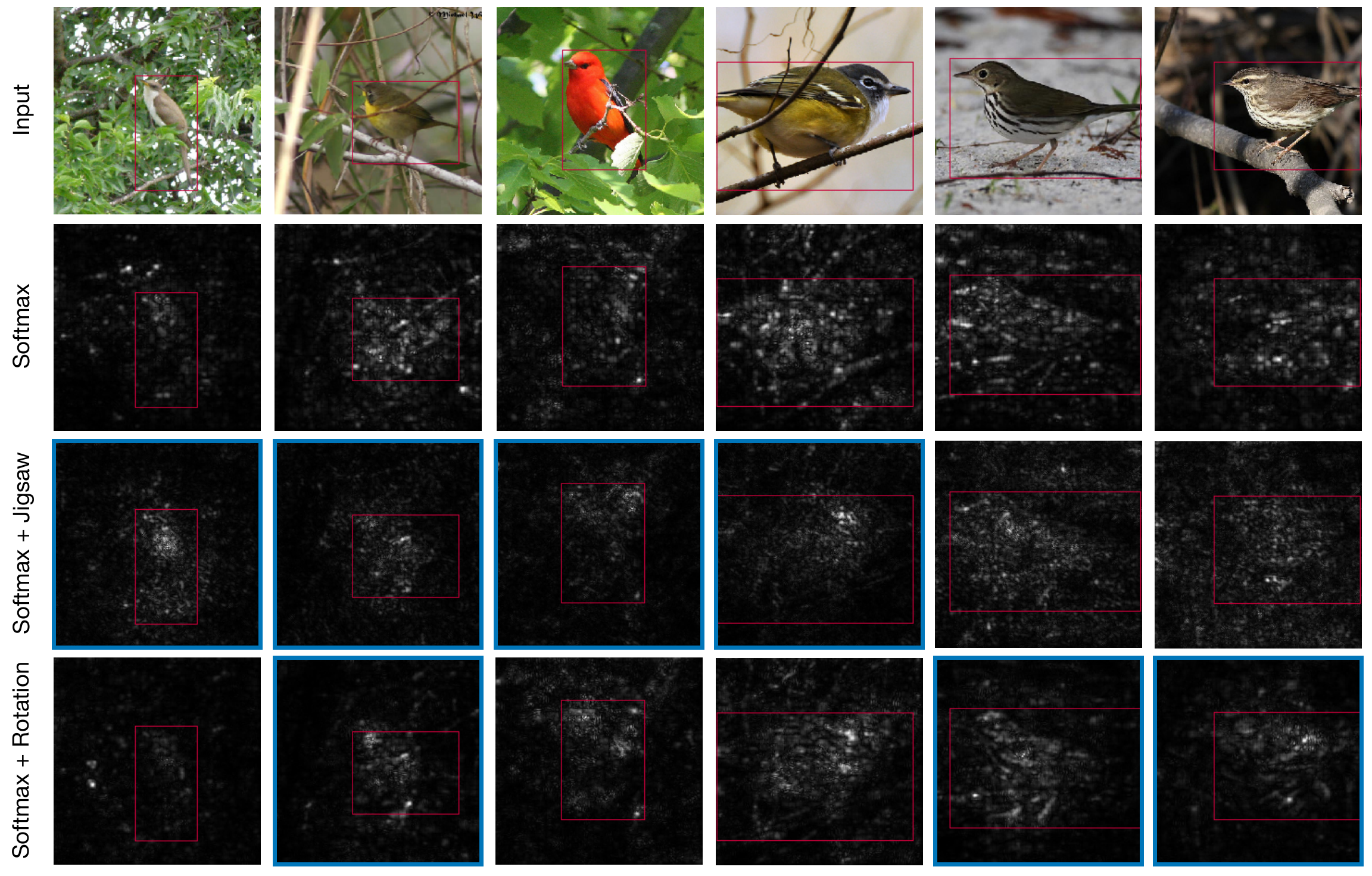}
\caption{\label{fig:saliency} \textbf{Saliency maps for various images
    and models.} For each image we visualize the magnitude of the
  gradient with respect to the correct class for models trained with
  various loss functions.
  The magnitudes are scaled to the same range for easier
  visualization.
  The models trained with self-supervision often have lower energy on
  the background regions when there is clutter. 
  We highlight a few examples with blue borders and the bounding-box of the object for each image is shown in red.
}
\end{figure}

To understand why the representation generalizes, 
we visualize what pixels contribute the most to the correct
classification for various models.
In particular, for each image and model, we compute the gradient of the
logits (predictions before softmax) for the correct class with
respect to the input image.
The magnitude of the gradient at each pixel is a proxy for its
importance and is visualized as ``saliency maps''.
Figure~\ref{fig:saliency} shows these maps for various images and
models trained with and without self-supervision on the standard classification task.
It appears that the self-supervised models tend to focus more on the
foreground regions, as seen by the amount of bright pixels within the
bounding box.
One hypothesis is that self-supervised tasks force the model to rely
less on background features, which might be accidentally correlated to
the class labels.
For fine-grained recognition, localization indeed improves performance
when training from few examples (see~\cite{WertheimerCVPR2019} for a contemporary
evaluation of the role of localization for few-shot learning).

\subsection{Experimental Details}\label{appendix:exp_details}
\subsubsection{Optimization details on few-shot learning}
During training, especially for the jigsaw puzzle task, we found it to be
beneficial to \emph{not} track the running mean and variance for the batch
normalization layer, and instead estimate them for each batch
independently. 
We hypothesize that this is because the inputs contain both full-sized
images and small patches, which might have different statistics. 
At test time we do the same. We found the accuracy goes up as the
batch size increases but saturates at a size of 64. 

When training with supervised and self-supervised loss, a trade-off term $\lambda$ between the losses can be used, thus the total loss is ${\cal L} = (1-\lambda){\cal L}_s + \lambda{\cal L}_{ss}$. We find that simply use $\lambda=0.5$ works the best, except for training on \textit{mini}- and \textit{tiered}-ImageNet with jigsaw loss, where we set $\lambda=0.3$. We suspect that this is because the variation of the image size and the categories are higher, making the self-supervision harder to train with limited data. When both jigsaw and rotation losses are used, we set $\lambda=0.5$ and the two self-supervised losses are averaged for ${\cal L}_{ss}$.

For training meta-learners, we use 16 query images per class for each training episode. When only 20\% of labeled data are used, 5 query images per class are used.
For MAML, we use 10 query images and the approximation method for backpropagation as proposed in~\cite{chen2019closerfewshot} to reduce the GPU memory usage. When training with self-supervised loss, it is added when computing the loss in the outer loop.
We use PyTorch~\cite{pytorch} for our experiments. 

\subsubsection{Optimization details on domain classifier}
For the domain classifier, we first obtain features from the penultimate-layer (2048 dimensional) from a ResNet-101 model pre-trained on ImageNet~\cite{ILSVRC15}. 
We then train a binary logistic regression model with weight decay using LBFGS for 1000 iterations. 
The images from the labeled dataset are the positive class and from the pool of unlabeled data are the negative class.
A subset of negative images are selected uniformly at random with 10 times the size of positive images.
A loss for the positive class is scaled by the inverse of its frequency to account for the significantly larger number of negative examples. 

\subsubsection{Optimization details on standard classification}
For standard classification (Appendix~\ref{appendix:fgvc}) we train a ResNet-18 network \emph{from scratch}. 
All the models are trained with ADAM optimizer with a learning rate of 0.001 for 600 epochs with a batch size of 16.
We track the running statistics for the batch normalization layer for the softmax baselines following
the conventional setting, \ie w/o self-supervised loss, but do not track these statistics when training with
self-supervision.

\subsubsection{Architectures for self-supervised tasks}
For jigsaw puzzle task, we follow the architecture of~\cite{noroozi2016unsupervised} where it was first proposed. 
The ResNet18 results in a 512-dimensional feature for each input, and we add a fully-connected
(\texttt{fc}) layer with 512-units on top. 
The nine patches give nine 512-dimensional feature vectors, which are concatenated. This
is followed by a \texttt{fc} layer, projecting the feature vector from
4608 to 4096 dimensions, and a \texttt{fc} layer with 35-dimensional outputs
corresponding to the 35 permutations for the jigsaw task.

For rotation prediction task, the 512-dimensional output of ResNet-18 is
passed through three \texttt{fc} layers with \{512, 128, 128, 4\}
units. The predictions correspond to the four rotation
angles. 
Between each \texttt{fc} layer, a \texttt{ReLU} activation and
a dropout layer with a dropout probability of 0.5 are added.

\end{document}